\documentclass[conference]{IEEEtran}

\IEEEoverridecommandlockouts

\usepackage[english]{babel}
\usepackage[T1]{fontenc}
\usepackage{amssymb, amsmath, amsthm}
\usepackage{mathtools}
\usepackage{cite}
\usepackage{graphicx}
\usepackage{subfloat}
\usepackage{subcaption} 
\usepackage{xcolor}
\usepackage{multirow}
\usepackage[ruled,vlined,linesnumbered]{algorithm2e}
\usepackage{enumitem}
\usepackage{stfloats}
\usepackage[normalem]{ulem}
\usepackage{stix}
\usepackage{booktabs}
\usepackage{tabularx}
\usepackage{multirow}
\usepackage{multicol}
\usepackage{amsmath}
\usepackage{algorithmic}
\usepackage{soul}

\usepackage{xcolor, colortbl}
\usepackage[acronym,,nohypertypes={acronym,notation}]{glossaries}

\definecolor{cian}{rgb}{.02,.7,.95}
\definecolor{gold}{rgb}{0.85,.66,0}

\title{A Primer on Kolmogorov-Arnold Networks (KANs) for Probabilistic Time Series Forecasting}

\author{
    Cristian J. Vaca-Rubio, \IEEEmembership{Member, IEEE}, Roberto Pereira, \IEEEmembership{Member, IEEE}, Luis Blanco, \IEEEmembership{Senior Member, IEEE}, \\ Engin Zeydan, \IEEEmembership{Senior Member, IEEE}, Màrius Caus, \IEEEmembership{Senior Member, IEEE} \\
    \thanks{Cristian J. Vaca-Rubio, Roberto Pereira, Luis Blanco, Engin Zeydan and Màrius Caus are with the Centre Tecnològic de Telecomunicacions de Catalunya (CTTC/CERCA), Castelldefels, Barcelona, Spain, 08860. (emails: \{cvaca, rpereira, lblanco, ezeydan, mcaus\}@cttc.es). This work has been submitted to IEEE for possible publication. Copyright may be transferred without notice, after which this version may not longer be available.}
}

\begin{document}
\maketitle

\begin{abstract}
This work introduces Probabilistic Kolmogorov-Arnold Network (P-KAN), a novel probabilistic extension of Kolmogorov-Arnold Networks (KANs) for time series forecasting. By replacing scalar weights with spline-based functional connections and directly parameterizing predictive distributions, P-KANs offer expressive yet parameter-efficient models capable of capturing nonlinear and heavy-tailed dynamics. We evaluate P-KANs on satellite traffic forecasting, where uncertainty-aware predictions enable dynamic thresholding for resource allocation. Results show that P-KANs consistently outperform Multi Layer Perceptron (MLP) baselines in both accuracy and calibration, achieving superior efficiency-risk trade-offs while using significantly fewer parameters. We build up P-KANs on two distributions, namely Gaussian and Student-$t$ distributions. The Gaussian variant provides robust, conservative forecasts suitable for safety-critical scenarios, whereas the Student-$t$ variant yields sharper distributions that improve efficiency under stable demand. These findings establish P-KANs as a powerful framework for probabilistic forecasting with direct applicability to satellite communications and other resource-constrained domains.
\end{abstract}

\begin{IEEEkeywords}
Probabilistic forecasting, Satellite, ML.
\end{IEEEkeywords}
\vspace{-0.3cm}
\section{Introduction}

Forecasting future network demand is essential for efficient resource allocation in wireless systems, particularly in satellites where spectral resources are scarce and traffic dynamics are highly nonlinear.  Accurate demand prediction enables proactive scheduling and energy-efficient operation, yet traditional deterministic models provide only point estimates and fail to capture the uncertainty inherent in traffic patterns. This limitation motivates the use of \textit{probabilistic forecasting}, in which models output a full predictive distribution rather than a single value, allowing the telecommunication network to balance efficiency and reliability according to risk preferences. Such forecasts have become central in domains such as energy, finance, and healthcare, where uncertainty-aware decisions are crucial~\cite{xie2024novel,cramer2022evaluation}. In wireless communications, and particularly in satellite systems, the ability to anticipate traffic variability with calibrated uncertainty estimates is equally critical: inaccurate forecasts may lead to over-provisioning (wasting scarce resources such as spectrum, power, or capacity) or under-provisioning (causing service degradation).

As in many other domains, recent advances in time series forecasting have been driven by deep learning. Transformer-based models have achieved strong performance by capturing long-range dependencies~\cite{zhou2021informer, zhang2023crossformer}, while large language models reprogrammed for time series tasks have shown robustness in data-scarce regimes~\cite{jin2023time}. However, these architectures are often parameter-heavy and computationally expensive, limiting their practicality in resource-constrained settings such as on-board satellite systems.

Therefore, in this work we focus on lightweight models. For instance, Kolmogorov-Arnold Networks (KANs) have emerged as a promising solution to challenge standard Multi-Layer Perceptron (MLP)-based architectures \cite{liuu2024kan}. Recent applications of KANs in time series forecasting have shown promising results \cite{vaca2024kolmogorov, zeydan2025f, dong2024kolmogorov}. Several authors have also attempted to adapt them to probabilistic frameworks for classification tasks \cite{hassan2024bayesian} and probabilistic forecasting \cite{jiang2025novel}. However, the probability insights of these works are ad-hoc solutions, applied after the model is already trained. In this letter, we introduce \emph{Probabilistic KANs (P-KANs)}, a novel framework that learns directly the parameters of predictive distributions. By embedding uncertainty modeling into the functional connections of KANs, our approach provides both expressive power and parameter efficiency. We evaluate P-KANs on satellite traffic forecasting, where predictive distributions support what we refer to as \textit{dynamic thresholding} for resource allocation. Results show that P-KANs consistently outperform MLP baselines in both accuracy and calibration, while requiring significantly fewer parameters. The Gaussian distribution provides robust forecasts suitable for safety-critical contexts, while the Student-$t$ variant produces sharper predictions that enhance efficiency under stable demand. These findings establish P-KANs as a promising framework for uncertainty-aware forecasting in satellite communications.
\vspace{-0.2cm}
\section{Motivation and Background}

In satellite systems, resource allocation varies over time with high volatility and occasional extremes, making point forecasts unreliable. Therefore, the goal is to generate predictive distributions that capture both expected values and uncertainty.

Formally, let ${y_t}_{t=1}^T$ denote the Physical Resource Block (PRB) allocation time series, representing the number of PRBs occupied at time $t$. Given a context window of $c$ past observations $y_{t-c:t-1} = (y_{t-c}, \ldots, y_{t-1})$, the task is to forecast the next $h$ future entries $y_{t:t+h-1} = (y_t, \ldots, y_{t+h-1})$. We seek to model the conditional joint distribution
\begin{equation}
    p(y_{t:t+h-1} \mid y_{t-c:t-1}; \Theta),
\end{equation}
where $\Theta$ are model parameters. To simplify optimization, we adopt the common assumption of conditional independence across forecast steps, factorizing the joint likelihood as
\begin{equation}
    p(y_{t:t+h-1} \mid y_{t-c:t-1}; \Theta) 
    = \prod_{i=0}^{h-1} p(y_{t+i} \mid y_{t-c:t-1}; \Theta).
\end{equation}

This formulation reduces the multi-step probabilistic forecasting problem to a sequence of per-step conditional likelihoods.  Crucially, by modeling predictive distributions of PRB demand, we obtain a direct proxy for constructing dynamic thresholds in satellite resource allocation: high predictive quantiles (e.g., Percentile 99) define provisioning levels that balance reliability with spectral efficiency, in contrast to the inefficiency of static maximum allocation.

\subsection{Likelihood-based Probabilistic Framework}

The likelihood formulation introduced above is general and can be combined with any neural architecture that outputs the parameters of a predictive distribution. In this work, we consider two Likelihood distributions\footnote{While this work focuses on Gaussian and Student-$t$ likelihoods, the framework can accommodate any parametric distribution.}: Gaussian and Student-$t$. Given the past context $\mathbf{y}_{t-c:t-1}$, in the first case, the model produces the mean $\mu_t$ and scale $\sigma_t$ of the distribution. For the Student-$t$ case, it also produces the degrees of freedom $\nu_t$. The conditional likelihood of the next observation $y_t$ can then be expressed under different distributional assumptions.

The Gaussian likelihood models light-tailed noise, while the Student-$t$ distribution is more robust to heavy-tailed behavior and outliers. This probabilistic framework is \textit{architecture-agnostic} and can be implemented using MLPs, KANs, Transformers, or any other neural network architecture. In this work, motivated by the limited computational and energy resources available for on-board satellite inference, we demonstrate its effectiveness using two simple and lightweight architectures: P-MLP and P-KAN.

\subsection{Probabilistic MLP (P-MLP)}

As a baseline, we adopt a Probabilistic Multilayer Perceptron (P-MLP). 
The past window $\mathbf{y}_{t-c:t-1}$ is mapped through a stack of dense layers, producing a hidden representation $\mathbf{h}_t \in \mathbb{R}^d$. 
From this representation, the predictive distribution parameters are obtained via linear heads: 
$\mu_t = \mathbf{W}_\mu^\top \mathbf{h}_t + b_\mu$, 
$\sigma_t = \mathrm{softplus}(\mathbf{W}_\sigma^\top \mathbf{h}_t + b_\sigma)$ and 
$\nu_t = 2 + \mathrm{softplus}(\mathbf{W}_\nu^\top \mathbf{h}_t + b_\nu)$.

Depending on the likelihood family, either $(\mu_t, \sigma_t)$ or $(\mu_t, \sigma_t, \nu_t)$ are used. P-MLP is conceptually simple and serves as a strong baseline, but it is limited by the fixed linear weight matrices of MLPs, which may not capture complex nonlinear structure in traffic dynamics.

\subsection{Probabilistic KAN}

To overcome these limitations, we propose a P-KAN.  
The design of P-KAN is motivated by the Kolmogorov-Arnold representation theorem, which states that any multivariate continuous function can be expressed as a finite superposition of univariate functions. Following this principle, each connection in a KAN layer is not a scalar weight, but a learnable univariate function. Concretely, a KAN layer computes the transformation of a given input $x_i \in \mathbb{R}$
\begin{equation}
    z_t = \sum_{i=1}^{n_\text{in}} \phi_{t,i}(x_i),
\end{equation}
where $n_{\text{in}}$ denotes the number of input dimensions of the layer and each connection function is parameterized as
\begin{equation}
    \phi(x) = w \, b(x) + s \sum_{r=1}^R c_r B_{k,r}(x).
\end{equation}
Here $b(x)$ is a smooth base activation (e.g., SiLU), $\{B_{k,r}(x)\}_{r=1}^R$ are B-spline basis functions of order $k$ and $w, s, \{c_r\}$ are learnable parameters. This formulation allows each connection to flexibly approximate nonlinear transformations, enabling the hidden state $\mathbf{z}_t$ to capture a more expressive functional structure than $\mathbf{h}_t$ in MLPs.

For probabilistic output, P-KAN manages differently to the MLP baseline.  
Instead of linear heads, each distribution parameter is predicted through a dedicated KAN layer: 
$\mu_t = f_\mu^{\text{KAN}}(\mathbf{z}_t)$, 
$\sigma_t = \mathrm{softplus}(f_\sigma^{\text{KAN}}(\mathbf{z}_t))$ and 
$\nu_t = 2 + \mathrm{softplus}(f_\nu^{\text{KAN}}(\mathbf{z}_t))$,  
where $f_\bullet^{\text{KAN}}$ denotes a spline-based mapping of the same form as (3).  
This dual use of KAN layers, in both hidden transformations and output heads, provides greater modeling flexibility to P-KANs than the fixed structure of P-MLP.

\subsection{Training Objective}

Training both P-MLP and P-KAN reduces to maximizing the conditional likelihood of the observed data.  
Equivalently, we minimize the negative log-likelihood (NLL) loss:
\begin{equation}
    \mathcal{L}(\Theta) 
    = - \sum_{t} \sum_{\tau=0}^{h-1} 
      \log p(y_{t+\tau} \mid \mathbf{y}_{t-c:t-1}; \Theta).
\end{equation}

\section{Dataset and Experimental Setup}

The dataset was generated as part of the European \textit{5G-STARDUST} project from real GEO satellite broadband traffic provided by a satellite operator \cite{5gstardust2024d41}. It contains hourly time series of Physical Resource Block (PRB) allocations across six beams, each covering over 500 anonymized users over one month. Although based on GEO data, the results are expected to generalize to LEO systems. According to (1), each model is conditioned on the past \(c=168\) hours (one week) and predicts the distribution for the next \(h=24\) hours (one day). For each beam, two weeks plus one day are used for training and one week plus one day for testing, with no overlap. Models are trained by minimizing the negative log-likelihood of the factorized conditional distribution in Eq.~(2). Finally, we compare probabilistic models with Point Forecast (PF) baselines to demonstrate the benefits of predictive uncertainty for satellite resource allocation, where misestimations directly impact spectral efficiency and service reliability. 
All models are trained for 300 epochs using Adam with a learning rate of \(10^{-3}\).

\vspace{-0.5 cm}
\section{Simulation results}

\subsection{Forecasting Results and Dynamic Thresholding}

We first demonstrate the forecasting performance of our proposed P-KAN model on the satellite resource allocation task and introduce the concept of \textit{dynamic thresholding}, where resource allocation decisions are driven by predictive uncertainty. Specifically, by using a high quantile of the forecast distribution (e.g., the P99 level), the allocation can adapt over time to traffic variability, offering a flexible alternative to static maximum provisioning and establishing a direct link between probabilistic forecasts and operational efficiency. This differs from traditional satellite system approaches that rely on fixed bandwidth, which cannot adapt to varying traffic demands and often result in inefficiencies \cite{pachler2020allocating}. Our approach moves away from this rigidity by leveraging probabilistic forecasts to enable dynamic allocation that better matches real-time conditions.

\begin{figure}[h!]
     \centering
     \begin{subfigure}[b]{0.45\textwidth}
        \includegraphics[width=\linewidth]{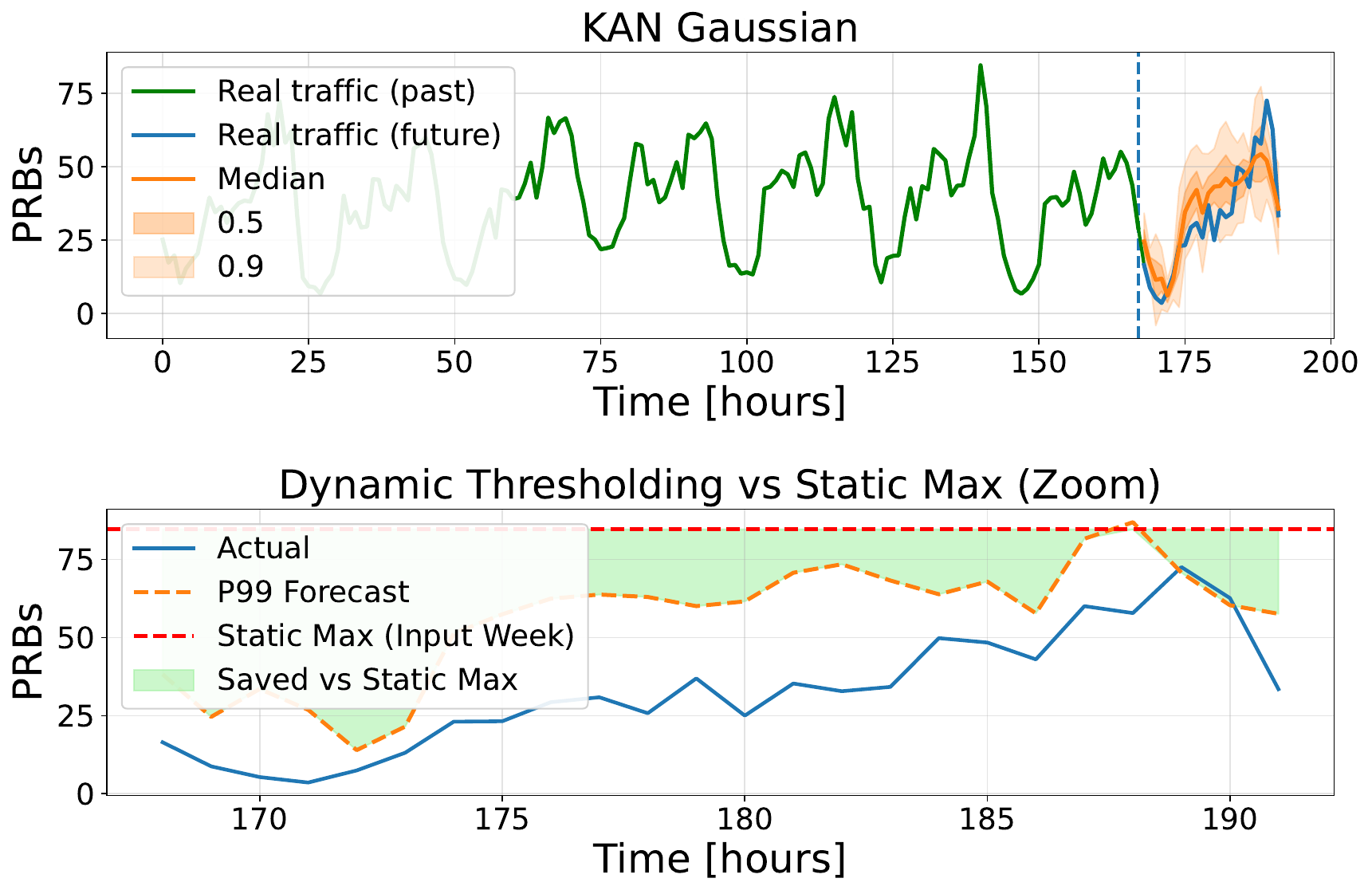}
    \caption{Gaussian likelihood}
    \label{fig:f_Plot_1}
     \end{subfigure}
     \begin{subfigure}[b]{0.45\textwidth}
         \centering
        \includegraphics[width=\linewidth]{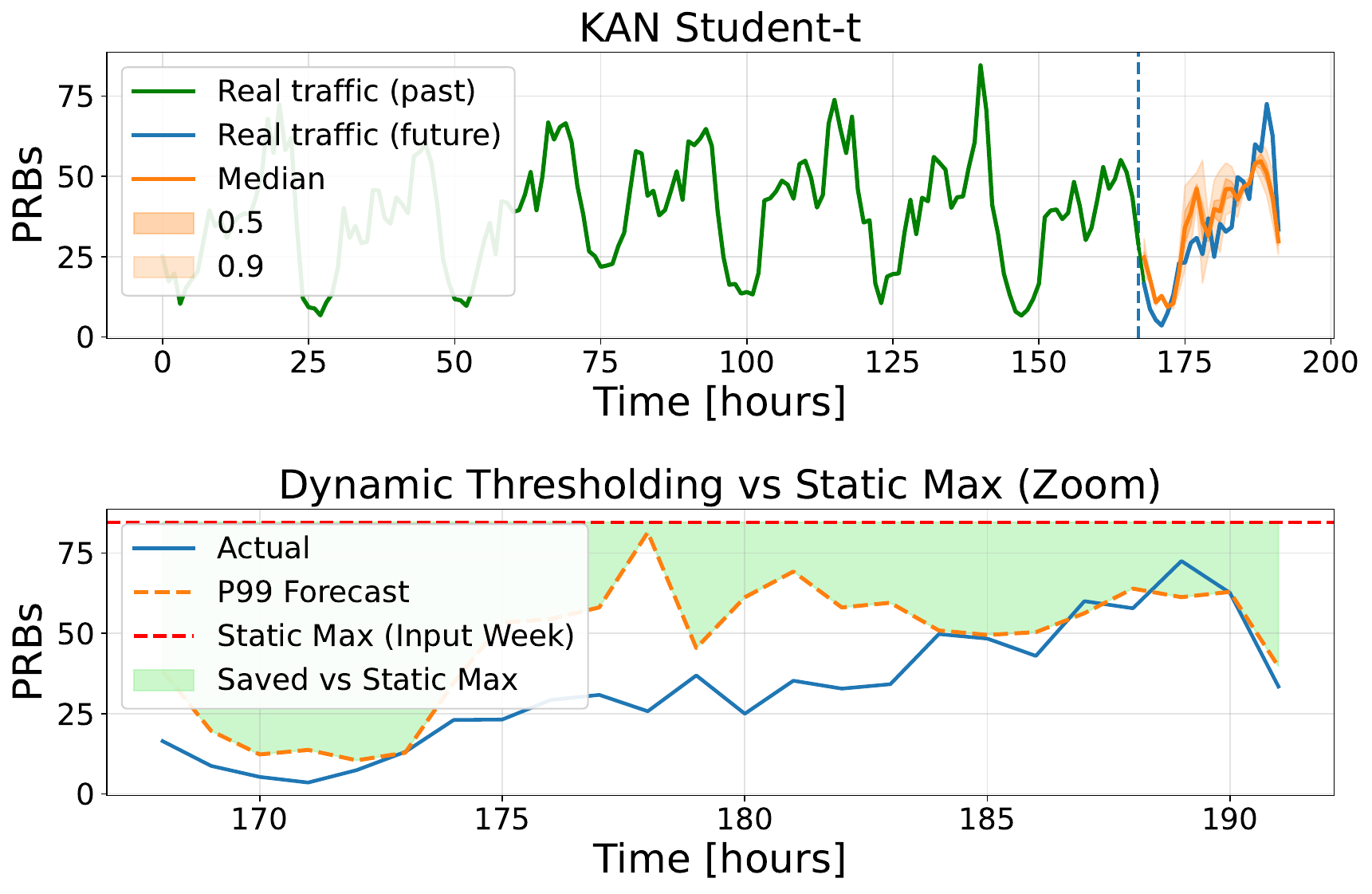}
       \caption{Student-$t$ likelihood}
        \label{fig:f_Plot_2}
     \end{subfigure}
\caption{Forecasts over one satellite beam using P-KAN  showing predictive intervals and adaptive thresholding (a) under Gaussian likelihood (b) under Student-$t$ likelihood.} 
\label{fig:forecast_plots}
\end{figure}

Figure~\ref{fig:forecast_plots} shows P-KAN forecasts under Gaussian and Student-$t$ likelihoods for the same beam. Both models capture the diurnal traffic patterns, but the width and behavior of their predictive intervals differ. The Gaussian likelihood produces relatively wider bands, resulting in more conservative predictions that allocate slightly more resources than necessary during stable periods. The Student-$t$ likelihood, by contrast, yields sharper central intervals and lower allocations under regular conditions, while still reacting to traffic bursts due to its heavy-tailed nature. 

The bottom panels illustrate the dynamic thresholding strategy, where allocation is determined adaptively using a high predictive quantile (P99) (dashed orange line) rather than a static maximum baseline (dashed red line). Both likelihoods achieve substantial savings (green area) compared to static allocation. However, Gaussian forecasts generally maintain the P99 threshold at a higher level for longer periods (for instance 180-185), ensuring that traffic surges are consistently covered and the risk of under-provisioning is minimised. Student-$t$ forecasts follow demand more closely and save slightly more resources (see section IV.B.3) during stable intervals, but at the expense of a higher probability of missing sudden peaks. Furthermore, the Gaussian likelihood offers the most robust trade-off: it delivers significant savings compared to static max while maintaining reliable coverage of extreme demand events. The Student-$t$ case is more efficient under normal conditions, but the Gaussian remains the safer choice for guaranteeing service continuity in operational settings.

\subsection{Quantitative Analysis}

We now present a detailed quantitative evaluation of the proposed probabilistic forecasting models. The analysis focuses on four aspects: (i) forecast accuracy, (ii) probabilistic accuracy, (iii) efficiency and risk trade-offs in resource allocation and (iv) model complexity.

\begin{table}[t]
\centering
\caption{Forecast accuracy comparison across models.}
\label{tab:forecast_accuracy}
\begin{tabular}{lccc}
\toprule
\textbf{Model} & \textbf{MSE} & \textbf{MAE} & \textbf{RMSE} \\
\midrule
KAN Gaussian     & 7.009  & 1.830 & 2.647 \\
\textbf{KAN Student-$t$}  & \textbf{4.347}  & \textbf{1.495} & 2.085 \\
MLP Gaussian     & 16.853 & 3.257 & 4.105 \\
MLP Student-$t$  & 21.015 & 3.182 & 4.584 \\
KAN-PF           & 4.532  & \textbf{1.495} & \textbf{1.951} \\
MLP-PF           & 5.144  & 1.580 & 2.056 \\
\bottomrule
\end{tabular}
\end{table}

\begin{table*}[t]
\centering
\caption{Probabilistic calibration metrics.}
\label{tab:calibration_metrics}
\resizebox{\textwidth}{!}{%
\begin{tabular}{lrrrrrrrrrr}
\toprule
\textbf{Model} & \textbf{CRPS} & \textbf{QL$_{0.1}$} & \textbf{QL$_{0.5}$} & \textbf{QL$_{0.9}$} & 
\textbf{Cov$_{0.1}$} & \textbf{Cov$_{0.5}$} & \textbf{Cov$_{0.9}$} & 
\textbf{FIC$_{0.1}$} & \textbf{FIC$_{0.5}$} & \textbf{FIC$_{0.9}$} \\
\midrule
KAN Gaussian    & 195.29 & \textbf{102.97} & 263.58 & \textbf{155.26} & \textbf{0.083} & \textbf{0.500} & \textbf{0.847} & \textbf{0.076} & \textbf{0.465} & \textbf{0.840} \\
KAN Student-$t$ & \textbf{184.92} & 113.20 & \textbf{215.34} & 203.75 & 0.347 & 0.472 & 0.597 & 0.014 & 0.132 & 0.333 \\
MLP Gaussian    & 407.81 & 396.70 & 469.06 & 304.13 & 0.569 & 0.694 & 0.792 & 0.014 & 0.111 & 0.306 \\
MLP Student-$t$ & 384.78 & 179.66 & 458.18 & 461.48 & 0.236 & 0.465 & 0.604 & 0.021 & 0.174 & 0.451 \\
\bottomrule
\end{tabular}%
}
\end{table*}

\subsubsection{Forecast Accuracy}
Forecasting accuracy is assessed through the Mean Squared Error (MSE), Mean Absolute Error (MAE) and Root Mean Squared error (RMSE), reported in Table~\ref{tab:forecast_accuracy}. These three measures highlight complementary aspects of error: MSE penalizes large deviations more heavily, MAE reflects the typical absolute deviation and RMSE provides a balanced indicator aligned with the traffic scale. For the probabilistic methods, these results are computed w.r.t. the median.
As shown in Table~\ref{tab:forecast_accuracy}, P-KAN models consistently outperform their P-MLP counterparts. The Student-$t$ variant of P-KAN achieves the lowest MSE and MAE, showing that it both suppresses extreme deviations and improves the typical forecast accuracy. The PF version, KAN-PF, attains the best RMSE, which underscores that spline-based functional connections capture short-term temporal dependencies more effectively than standard dense layers, even without probabilistic modeling. In contrast, MLP-based models yield substantially higher errors across all three metrics, with RMSE values more than twice those of the best P-KAN variants. This indicates that fixed-weight connections are less expressive when modeling the bursty and highly variable nature of satellite traffic. In practice, the superior accuracy of P-KAN translates into forecasts that are better aligned with the true demand, reducing the mismatch that drives inefficient resource allocation in satellite systems.

\begin{figure}[t!]
\centering
\includegraphics[width= 0.9\linewidth]{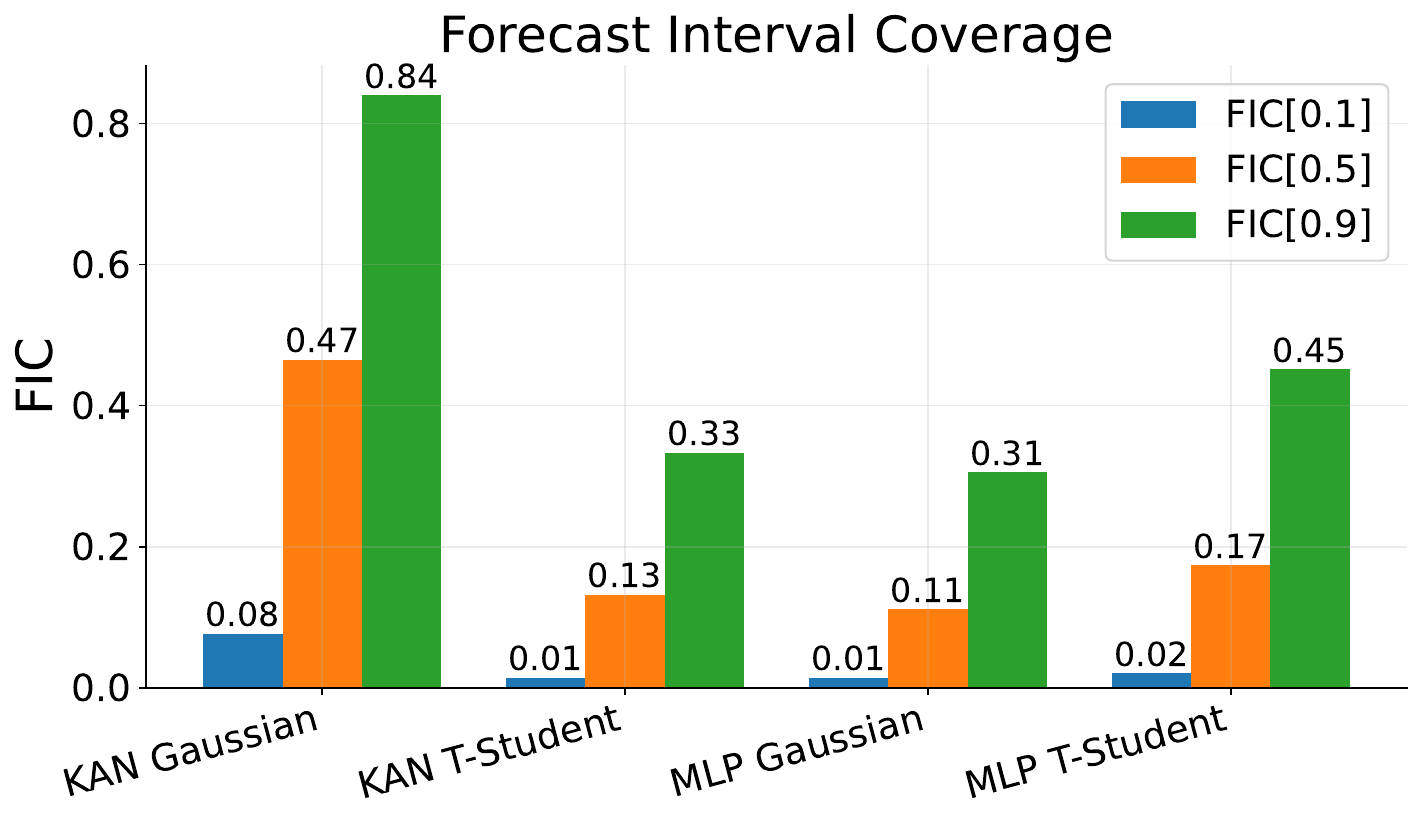}
\caption{FIC at the 10\%, 50\% and 90\% nominal levels 
for probabilistic models. }

\label{fig:fic}
\end{figure}

\subsubsection{Probabilistic Accuracy}

We next evaluate the ability of the models to produce well-calibrated predictive distributions. We use standard metrics: the Continuous Ranked Probability Score (CRPS) measures the overall calibration and sharpness of the predictive distributions; the Quantile Loss (QL) assesses the accuracy of specific predictive quantiles; and Coverage and Forecast Interval Coverage (FIC), which quantify the empirical consistency between predicted probability intervals and observed outcomes.

Figure~\ref{fig:fic} shows the FIC at nominal levels $0.1$, $0.5$ and $0.9$. The KAN Gaussian model achieves coverage values closest to the nominal targets, demonstrating the most reliable calibration across intervals. In contrast, the Student-$t$ variants yield sharper but less conservative intervals, often under-covering extreme demand variations. MLP-based models perform noticeably worse, with FIC deviating substantially from the desired levels. This result highlights the benefit of spline-based functional connections in P-KAN, which better balance sharpness and calibration than fixed-weight MLPs.

Table~\ref{tab:calibration_metrics} presents a more detailed quantitative analysis of probabilistic accuracy. The CRPS further confirms the superiority of P-KAN models, with the Student-$t$ variant achieving the lowest CRPS, indicating sharper and more informative forecasts. However, this improvement comes at the cost of reduced coverage consistency, as reflected by both coverage and FIC. In contrast, the KAN Gaussian maintains a stronger calibration trade-off, showing both competitive CRPS and the most accurate interval coverage among all models. MLP baselines exhibit significantly higher CRPS and quantile losses, confirming their limited ability to capture distributional uncertainty. These results demonstrate that P-KAN models not only improve point-wise forecast accuracy but also enhance the probabilistic characterisation of uncertainty. The Gaussian variant offers robust and conservative forecasts suitable for safety-critical resource allocation, while the Student-$t$ variant provides sharper distributions that may yield higher efficiency in stable conditions but at the expense of increased risk under traffic surges.

\begin{figure}[t!]
\centering
\includegraphics[width= 0.9\linewidth]{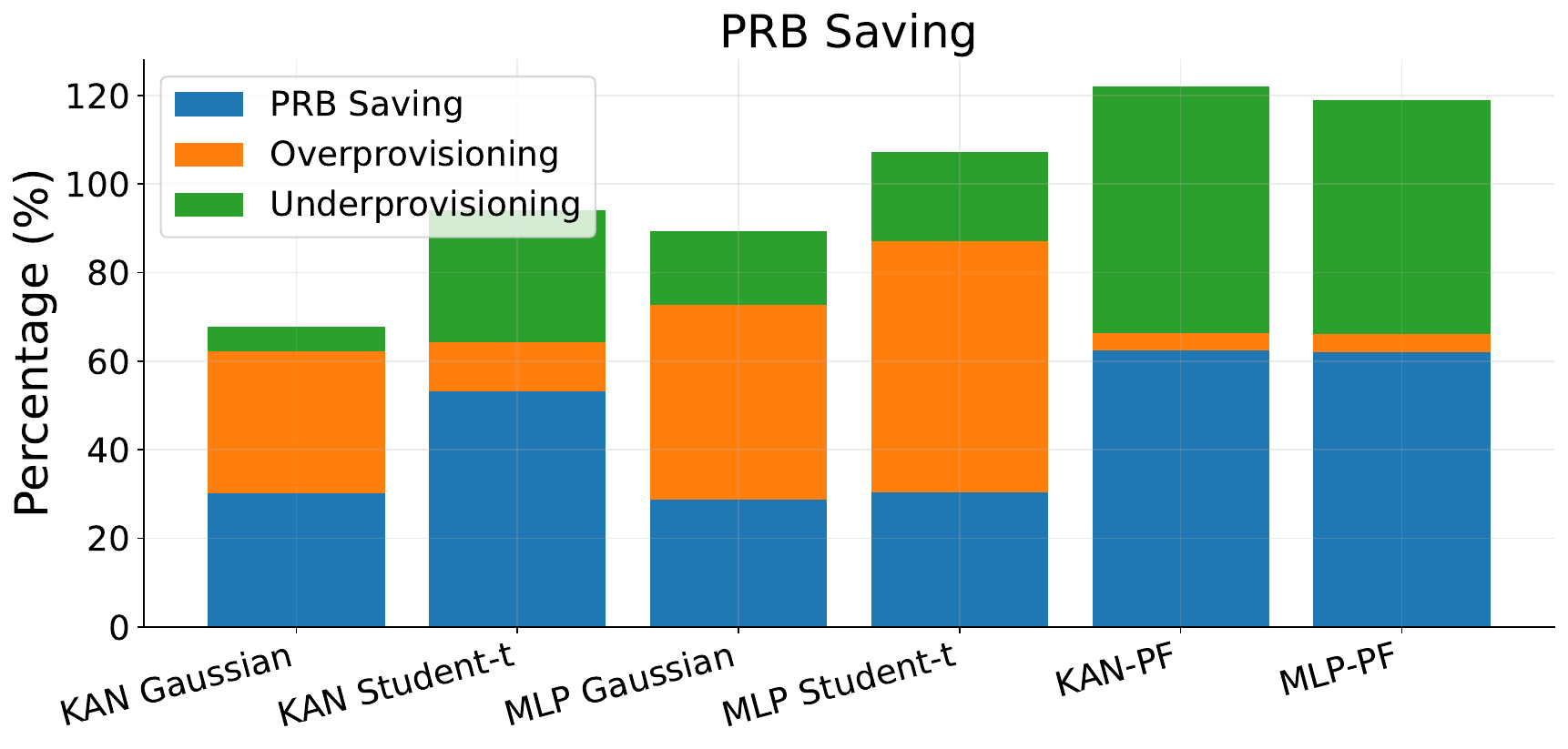}
\caption{Breakdown of resource allocation efficiency and risk across models, showing the proportion of PRB savings, overprovisioning and underprovisioning. }

\label{fig:tradeoff}
\end{figure}

\begin{figure}[t!]
\centering
\includegraphics[width= 0.9\linewidth]{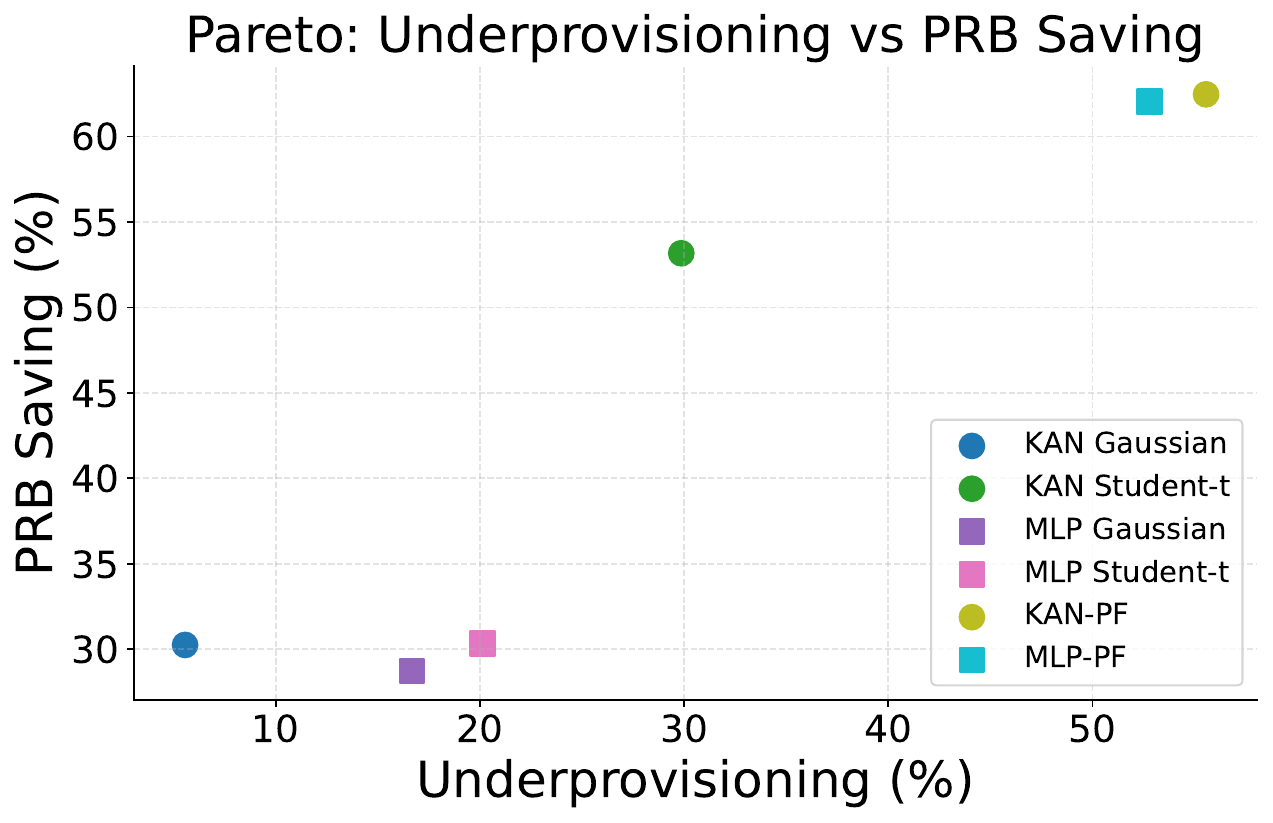}
\caption{Pareto analysis of PRB savings versus underprovisioning. }

\label{fig:pareto}
\end{figure}

\subsubsection{Efficiency and Risk Trade-offs}

Beyond introducing the novel P-KAN framework, another key contribution of this work is its practical relevance to satellite resource allocation. In this context, efficiency is quantified as the percentage of PRBs saved relative to the static-max baseline, while risk is measured by the rates of overprovisioning and underprovisioning. These conclusions are based on the use of P99 dynamic thresholding. Figures~\ref{fig:tradeoff} and \ref{fig:pareto} summarise the main trade-offs across models. Figure~\ref{fig:tradeoff} presents a stacked decomposition of PRB savings, overprovisioning, and underprovisioning. The proposed P-KAN models consistently outperform the MLP baselines, achieving greater PRB savings with more favourable provisioning balances. Among these, the P-KAN Student-$t$ variant achieves the highest savings (over 50\%), but at the cost of increased underprovisioning. In contrast, the P-KAN Gaussian variant is more conservative, providing around 30\% savings while keeping underprovisioning to a minimum, making it more robust in safety-critical contexts. The MLP models lag behind in both efficiency and provisioning reliability, highlighting the advantages of spline-based functional connections.

These trade-offs are further clarified in Figure~\ref{fig:pareto}, which plots PRB savings against underprovisioning. The frontier reveals two distinct operating points: the Student-$t$ variant, which favours efficiency and aggressive savings, and the Gaussian variant, which prioritises robustness with lower risk. PF baselines mark an upper bound in PRB savings but incur excessive underprovisioning, making them impractical for real-world deployments. These results emphasise that the choice of likelihood is not merely a modelling detail but determines the system-level behavior of P-KAN forecasts. In particular, Gaussian-based models are better suited when service continuity is critical, while Student-$t$ models may be preferable in efficiency-driven scenarios where occasional underprovisioning is acceptable.

\begin{figure}[t!]
\centering
\includegraphics[width=0.9\linewidth]{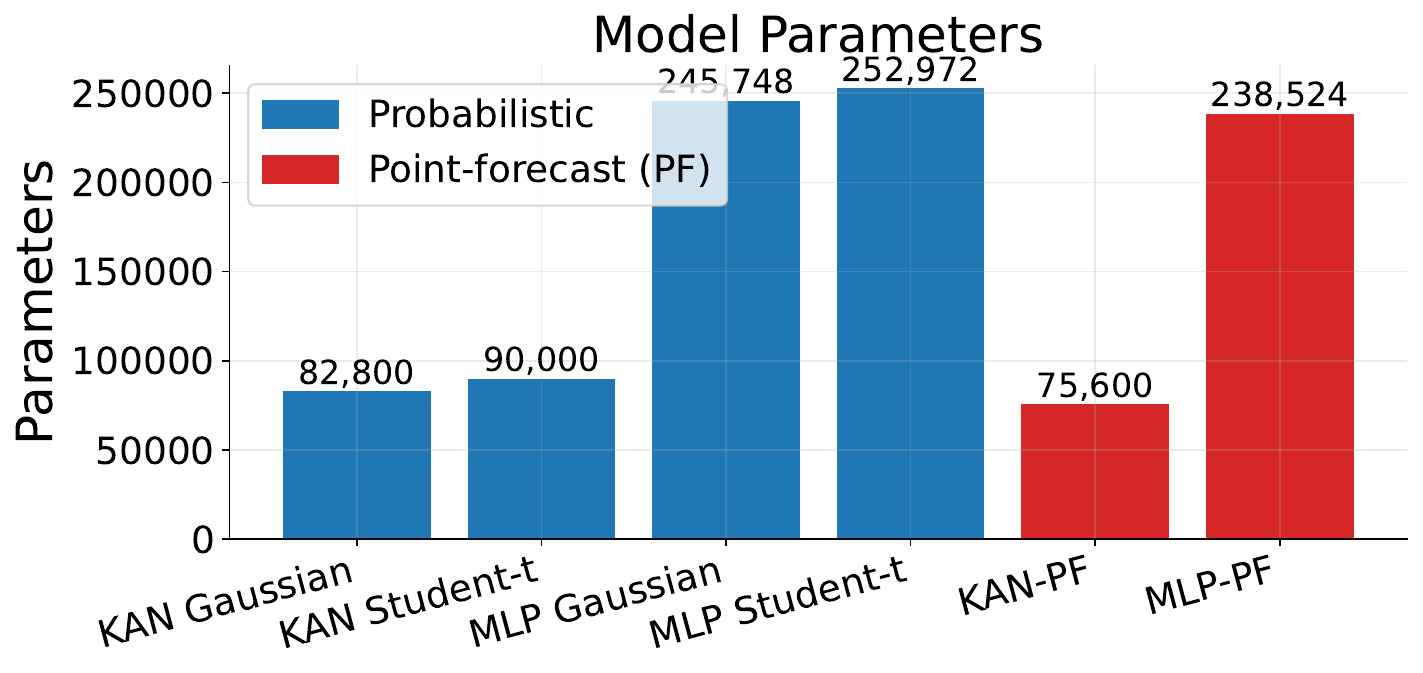}
\caption{Trainable parameters across probabilistic and PF models. }
\label{fig:params}
\end{figure}

\subsubsection{Model Complexity}

Finally, Fig.~\ref{fig:params} compares the number of trainable parameters across models. Despite delivering superior forecasting and allocation performance, P-KAN models require significantly fewer parameters (\(\approx 82\text{k} - 90\text{k}\)) than MLP baselines (\(>240\text{k}\)). This parameter efficiency is particularly relevant in the satellite domain, where on-board compute, memory, and energy resources are severely constrained. By encoding richer transformations per weight through spline-parameterised functional connections, P-KAN reduces redundancy while preserving expressiveness. This makes the proposed architecture not only more accurate but also more practical for deployment in real satellite systems, enabling efficient in-orbit resource management without exceeding hardware budgets. Moreover, fewer parameters also imply faster inference, which is crucial for real-time allocation decisions on board. 

\vspace{-0.5 cm}
\section{Conclusion}

This letter presented P-KANs, a probabilistic extension of KANs for time series forecasting. By replacing scalar weights with spline-based functional connections, P-KANs jointly improve predictive accuracy and uncertainty modelling. Evaluated on real satellite traffic data, they outperform MLP baselines in calibration and efficiency-risk trade-offs while requiring far fewer parameters. Gaussian and Student-$t$ variants balance robustness and efficiency, making P-KANs a lightweight solution for on-board resource management.
\section*{Acknowledgment}
This work was supported by the European Union’s Horizon Europe Smart Networks and Services Joint Undertaking (SNS JU) under the 5G-STARDUST project (Grant No. 101096573) and the UNITY-6G project (Grant No. 101192650), as well as by the SOFIA project (PID2023-147305OB-C32, MICIU/AEI/FEDER, UE) and the CHIST-ERA SONATA project (CHIST-ERA-20-SICT-004, PCI2021-122043-2A).

\bibliographystyle{IEEEtran}
\bibliography{reference.bib}

\end{document}